\documentclass[10pt, conference]{IEEEtran}
\IEEEoverridecommandlockouts
\usepackage{cite}
\usepackage{amsmath,amssymb,amsfonts}
\usepackage{algorithmic}
\usepackage{graphicx}
\usepackage{textcomp}
\usepackage{multirow}
\usepackage{hyperref}
\usepackage{xcolor}
\def\BibTeX{{\rm B\kern-.05em{\sc i\kern-.025em b}\kern-.08em
    T\kern-.1667em\lower.7ex\hbox{E}\kern-.125emX}}
\begin{document}

\title{Extracting Usable Predictions from Quantized Networks through Uncertainty Quantification for OOD Detection\\
}

\author{\IEEEauthorblockN{Rishi Singhal}
\IEEEauthorblockA{\textit{Dept of Computer Science} \\
\textit{North Carolina State University}\\
Raleigh, USA \\
rsingha4@ncsu.edu}
\and
\IEEEauthorblockN{Srinath Srinivasan}
\IEEEauthorblockA{\textit{Dept of Computer Science} \\
\textit{North Carolina State University}\\
Raleigh, USA \\
ssrini27@ncsu.edu}
}

\maketitle

\begin{abstract}
OOD detection has become more pertinent with advances in network design and increased task complexity. Identifying which parts of the data a given network is misclassifying has become as valuable as the network's overall performance. We can compress the model with quantization, but it suffers minor performance loss. The loss of performance further necessitates the need to derive the confidence estimate of the network's predictions. In line with this thinking, we introduce an Uncertainty Quantification(UQ) technique to quantify the uncertainty in the predictions from a pre-trained vision model. We subsequently leverage this information to extract valuable predictions while ignoring the non-confident predictions. We observe that our technique saves up to 80\% of ignored samples from being misclassified. The code for the same is available \href{https://github.com/rishi2019194/Deep_Learning_Project}{here}
\end{abstract}

\section{Introduction}

\begin{figure}
    \centering
    \includegraphics[width=1.2\linewidth]{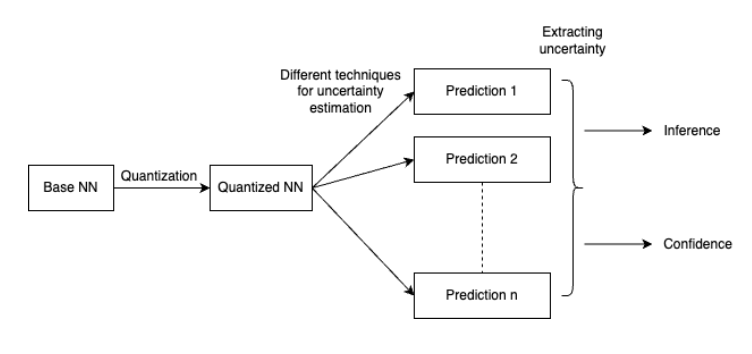}
    \caption{Representative Figure of the proposed UQ technique}
    \label{Representative Figure}
\end{figure}

Performing out-of-distribution detection is useful when we are interested in finding samples the model is not designed to categorize and thus might not perform well. Finding such samples allows us to avoid misclassifying and wasting manual human resources to find which parts of the dataset the model has made a mistake on. In safety-critical scenarios such as autonomous driving, OOD detection could be a great way to decide when human intervention is necessary \cite{genOOD}. Hence. in the pursuit of improving OOD detection, our work introduces an Uncertainty Quantification(UQ) technique to enhance the OOD detection process in deep learning models, especially in resource-constrained environment, as shown in Figure \ref{Representative Figure}.\\

Our contributions are as follows:
\begin{itemize}
    \item We first introduce Monte-Carlo dropouts into a pre-trained, model fine-tuned on the last few layers followed by post-training integer quantization. We collect slightly different outputs from the model we receive from performing Monte-Carlo dropouts over several iterations. 
    \item We then quantify the uncertainty of the model from these predictions by calculating a confidence interval within which we expect the accurate prediction to fall. 
    \item We then leverage this uncertainty to filter better predictions while ignoring the rest. 
    \item We analyze the CIFAR-100\cite{cifar100} and further validate it on the corrupted CIFAR-100C\cite{cifar100c} dataset.
\end{itemize}

\section{Related Works}
Several papers have previously researched the OOD Detection and Uncertainty Quantification field, some of which are mentioned below.

Liu et al. \cite{liu2020energy} proposed a unified framework for OOD detection by replacing the softmax score with a free-energy-based score. Compared to the softmax scores, the proposed metric is less susceptible to overconfident posterior distributions for OOD data samples. They observe that the energy-scoring function performs well during both inference using pre-trained models and even for training from scratch for OOD Detection. The proposed energy-based metrics helped achieve state-of-the-art results by training existing OOD detection models.

Yang et al.\cite{yang2023mixood} tried tackling the problem of adding extra auxiliary heterogeneous datasets in the training stage for OOD Detection. However, existing papers did not ensure that the auxiliary dataset added was of high quality, i.e., OOD samples are close to the In-distribution dataset but are still classified as OOD. Thus, to combat this problem, a Mixup-based OOD detection strategy was proposed to generate augmented images and test the same on existing OOD models like Maximum Softmax Probability\cite{hendrycks2016baseline}, Energy Score\cite{liu2020energy}, and ODIN\cite{liang2017enhancing}.

Sun et al.\cite{sun2022dice} focused on the idea that existing methods for OOD Detection are highly parameterized and try to prune unimportant weights and units attributed to the brittleness of OOD detection. Thus, they applied a sparsification methodology(DICE) for ranking network weights using a measure of contribution and using those important weights for OOD Detection.

Lin et al.\cite{lin2021mood} exploited intermediate network layer outputs to develop multi-level OOD detection classifiers. They tried using intermediate OOD detectors at varying network depths to enable dynamic inference. They also try to compute the optimal exit point for a given test sample in the forward pass by an image compression-bits technique. Lastly, they also proposed an adjusted energy score for OOD detection, tested their proposed technique compared to existing models, and got better results(accuracy, AUCROC, FLOPs requirements).

Lastly, Gal et al.\cite{gal} forms the basis for all works that introduce inference time perturbation techniques to model the uncertainty in the model's predictions. They substantiate with mathematical proof that network outputs through inference time dropouts are similar to outputs from Bayesian Neural Networks(BNN) whose weights are samples from distributions.

\section{Methodology}
This section presents our proposed UQ technique, which can be applied to different machine learning and deep learning models. We first discuss the quantization aspect of the methodology, followed by the inference time dropout, which forms the basis of the UQ technique. Lastly, we describe how we utilize the output of the UQ technique to determine whether a sample is an OOD sample or not.\\

In the initial phase, we use two different pre-trained models - ResNet50\cite{resnet} and EfficientNet-B0\cite{efnet} and fine-tune them on the CIFAR-100 dataset. Now, on the trained model, we apply post-training full integer(int8) quantization present in the Tensorflow framework\cite{TensorFlow_Quant}. By doing so, we convert all weights and activation outputs into 8-bit integer data, i.e., 0-255. We then use this quantized model to do the inference.

Similar to dropout\cite{dropout} during the inference time, we randomly ignore a certain number of weights based on the dropout ratio and calculate the output. This procedure is repeated multiple times to generate slightly different outputs each time, as different weights are ignored during each pass. We assume that these multiple predictions are from a Gaussian, as evidenced by Gal et al. \cite{gal}.

Once the multiple different outputs are collected through the MC dropout technique, the per-class average and standard deviation are calculated. The confidence interval is then calculated as described in (\ref{uint})

\begin{equation}
\label{uint}
    \left ( \mu -Z\sigma , \mu +Z\sigma \right )
\end{equation}

where \(\mu\) is the per class mean of the multiple outputs, \(\sigma\) the standard deviation, and \(Z\) the \(z\)-table value of the percentage confidence interval which we term as the conf\_factor.

Once we have calculated the confidence interval for each class among the multiple outputs, we compare them to our set threshold. If the entire interval lies above the threshold, we assign the class as present or 1. If the entire interval lies below the threshold, we assign the class as absent or 0. If the interval lies between the threshold, our model is uncertain about this class's presence, so we assign it a -1 value.

If our final prediction vector now contains a -1 but no 1s, we ignore this prediction. If the prediction contains no -1s but contains 1s, we keep this prediction. If the prediction is all 0s, we again ignore the prediction. If we find the prediction to have a single -1 but all other 0s, we decide to give the benefit of the doubt to the model and mark the -1 as 1. This means that the model's prediction has fallen right around the threshold value, indicating that it is half sure that the particular class is present.

\section{Experiments \& Results}
In this section, we explain how we set up and fine-tune pre-trained models to detect OOD samples and demonstrate the experimental results of the proposed method.

\subsection{Experimental Setup}

\subsubsection{CIFAR-100 Dataset}
We fine-tuned the pre-trained models on the CIFAR-100 dataset, a standard benchmark dataset for image classification and OOD Detection. The CIFAR-100 dataset consists of 60000 32x32 colored images with 100 "fine labels (the class to which the sample belongs)." Therefore, there are 500 training images and 100 testing images per class. For our experimental setup, we used all 50000 samples for training and 100 samples for testing.\\

\subsubsection{CIFAR-100C Dataset}
For the case of OOD samples, we used the CIFAR-100-Corrupted(CIFAR-100C) dataset. The CIFAR-100C dataset consists of 60000 32x32 colored images for each type of noise (19 noises) with two different severities - severity 1 and 5. We used the same for inference purposes only, i.e., to test the fine-tuned models on OOD samples. For our experimental setup, we used 20 samples for each type of noise and severity.\\

\subsubsection{Implementation Details}
\begin{figure*}[htbp]
\centerline{\includegraphics[width=\textwidth]{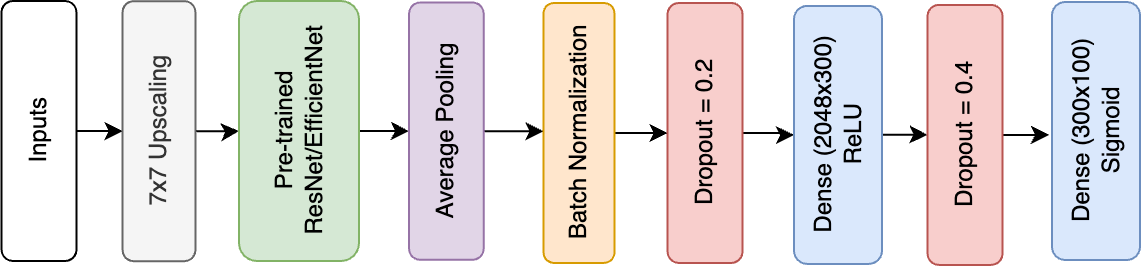}}
\caption{Model architecure}
\label{model}
\end{figure*}

For our model(Fig \ref{model}), we adopted the ResNet50 and EfficientNet-B0 pre-trained models as our primary classification network. We fine-tuned both of the two networks by replacing the classification layer with Average Pooling followed by Batch Normalization, a Dropout of 0.2, a Dense layer(2048x300) with ReLU activation, a Dropout of 0.4, and another Dense layer(300x100) with sigmoid activation function. Apart from that, we used hyperparameters of 10 epochs, Adam\cite{adam} optimizer, Binary Cross Entropy Loss, and a learning rate of 0.01.

\subsubsection{Evaluation Metrics}
We evaluated our model's performance using two different metrics - F1-Score and Misclassified(\%).

\textbf{Micro Average F1-Score:}
Since we have a multi-label dataset, we used the micro average F1-score for evaluation. It makes use of - Net True Positive(TP), Net False Positive(FP), and Net False Negative(FN) values.

Here, Net TP refers to the sum of class-wise TP scores, Net FP refers to the sum of class-wise FP scores, and Net FN refers to the sum of class-wise FN scores of the dataset. The micro-F1-Score metric is as described in (\ref{microf1}).

\begin{equation}
\label{microf1}
Micro-F1 = \frac{Net\;TP}{Net\;TP + \frac{1}{2}(Net\;FP + Net\;FN)}
\end{equation}

Also, since CIFAR-100 and CIFAR-100C are binary label datasets, hence the micro-F1-score simply becomes accuracy.

\textbf{Ignored Samples:}
Since our task is concerned with detecting OOD samples and discarding them due to their confusing nature, we define another performance metric of the number of ignored samples(I), which the quantized model results in after following the UQ technique.

\textbf{Misclassified(\%):}
According to the proposed UQ technique, we ignore specific samples whose threshold lies within the confidence interval, i.e., the model is uncertain which class the sample belongs to. Hence, to observe how much \% of these ignored samples are getting misclassified by the original quantized model, we proposed a Misclassified(\%) metric as described in (\ref{misclassified}).

\begin{equation}
\label{misclassified}
Misclassified(\%) = \frac{Number\;of\;MI\;samples}{Number\;of\;I\;samples}
\end{equation}

where MI refers to Misclassified-Ignored samples predicted by the original quantized model, and I refers to the Ignored samples predicted by our UQ technique.

\subsection{Experiments}
\subsubsection{CIFAR-100 Experiments}
At first, we fine-tuned the ResNet50 and EfficientNet-B0 models using the CIFAR-100 dataset by following the model architecture as explained before and proceed with quantization and UQ technique.\\

\textbf{Training \& Int8 Quantization}\\
 After that, we carried out post-training full integer(int8) quantization to get a reduced model size \~4x times, as described in Table \ref{quantizationsize}. Finally, we carried out inference on 100 test samples of the CIFAR-100 dataset without the proposed UQ technique on the original quantized model. We compared it with the non-quantized model setup as seen in Table \ref{cifar100test}.

\begin{table}[htbp]
\caption{Model size before and after int8 quantization}
\begin{center}
\begin{tabular}{|c|c|c|}
\hline
\textbf{Model Type} & \textbf{Unquantized Model} & \textbf{Quantized Model} \\
\hline
\textbf{ResNet50} & 92.1MB & 23.7MB \\
\hline
\textbf{EfficientNet B0} & 92.2MB & 23.6MB \\
\hline
\end{tabular}
\label{quantizationsize}
\end{center}
\end{table}

\begin{table}[htbp]
\caption{F1 scores before and after quantization}
\begin{center}
\begin{tabular}{|c|c|c|}
\hline
\textbf{Model Type} & \textbf{Unquantized Model} & \textbf{Quantized Model} \\
\hline
\textbf{ResNet50} & 73\% & 72\% \\
\hline
\textbf{EfficientNet B0} & 73\% & 71\% \\
\hline
\end{tabular}
\label{cifar100test}
\end{center}
\end{table}

\textbf{UQ-Technique: Grid Search CV}\\
Now, we use our proposed UQ technique to detect and ignore OOD samples. To do so, we first did a grid-search CV by varying the confidence factor(conf\_factor) and the number of inference iterations(num\_iter) by using the ResNet50 model, which is essentially described in Tables \ref{gridsearchf1} and \ref{gridsearch_ignored}. From this, we could find a trade-off between the num\_iter and conf\_factor values while maintaining sufficiently good F1-Score and number of ignored samples. 

\begin{table}[htbp]
\caption{F1 score in each configuration after performing Grid Search CV on ResNet50}
\begin{center}
\begin{tabular}{|c|c|c|c|}
\hline
\textbf{F1-Score} & \textbf{Num\_iter =} & \textbf{Num\_iter =} & \textbf{Num\_iter =} \\
& \textbf{20} & \textbf{30} & \textbf{50} \\
\hline
\textbf{Conf\_factor = 0.7} & 86.48 & 84.21 & 85.71 \\
\hline
\textbf{Conf\_factor = 0.8} & 85.52 & 86.84 & 86.11 \\
\hline
\textbf{Conf\_factor = 0.9} & 92.06 & 88.23 & 86.3 \\
\hline
\end{tabular}
\label{gridsearchf1}
\end{center}
\end{table}

\begin{table}[htbp]
\caption{Number of samples ignored in each configuration after performing Grid Search CV on ResNet50}
\begin{center}
\begin{tabular}{|c|c|c|c|}
\hline
\textbf{\# of Ignored} & \textbf{Num\_iter =} & \textbf{Num\_iter =} & \textbf{Num\_iter =} \\
\textbf{Samples} & \textbf{20} & \textbf{30} & \textbf{50} \\
\hline
\textbf{Conf\_factor = 0.7} & 26 & 24 & 23 \\
\hline
\textbf{Conf\_factor = 0.8} & 24 & 24 & 28 \\
\hline
\textbf{Conf\_factor = 0.9} & 37 & 32 & 27 \\
\hline
\end{tabular}
\label{gridsearch_ignored}
\end{center}
\end{table}

We observe that conf\_factor of 0.7 and num\_iter of 50 iterations seem to result in a good enough F1-score and number of ignored sample values, and so we picked these hyper-parameters for further experiments with ResNet50 and EfficientNet-B0.\\

\textbf{Visual Analysis for ResNet50 \& EfficientNet-B0}\\
Apart from the F1-score and the number of ignored samples, we also did a visual analysis of the samples that were ignored by the quantized model by following the UQ technique.

\begin{figure}[h]
  \includegraphics[width=.15\textwidth]{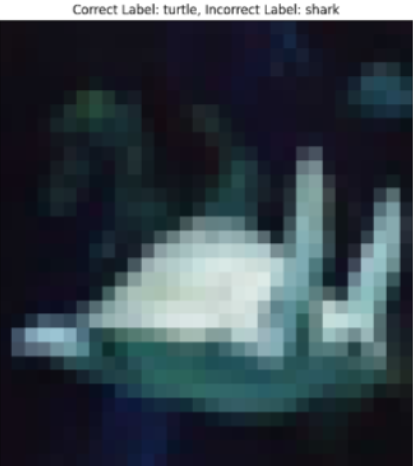}  
  \includegraphics[width=.15\textwidth]{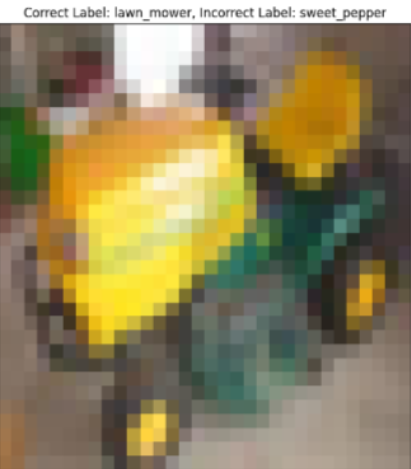}
  \includegraphics[width=.15\textwidth]{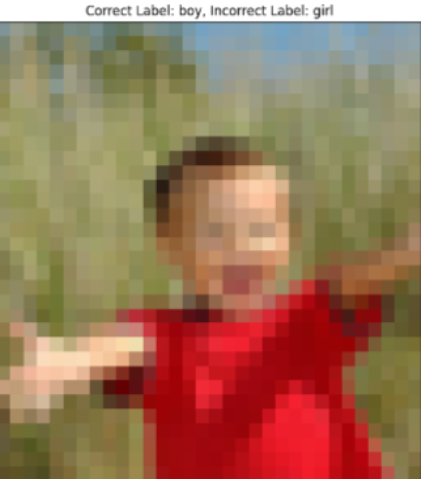}
  \caption{Some images ignored by ResNet50 a) Correct: turtle, Predicted: shark; b) Correct: lawn\_Mover, Predicted: sweet\_pepper; c) Correct: boy, Predicted: girl}
  \label{igr}
\end{figure}

\begin{figure}[h]
  \includegraphics[width=.15\textwidth]{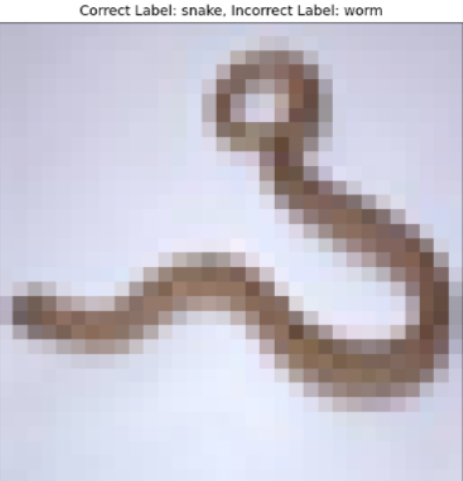}  
  \includegraphics[width=.15\textwidth]{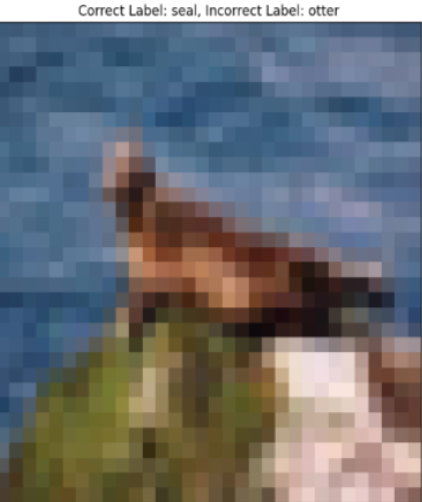}
  \includegraphics[width=.15\textwidth]{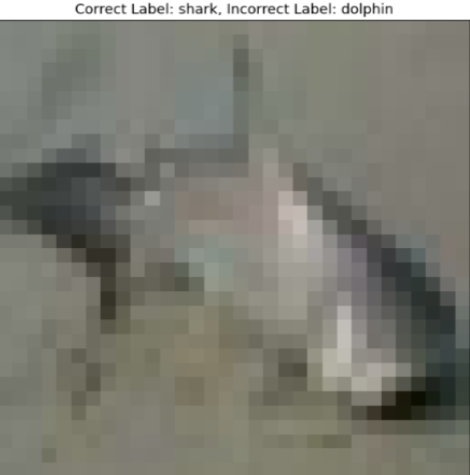}
  \caption{Some images ignored by EfficientNetB0 a) Correct: snake, Predicted: worm; b) Correct: seal, Predicted: otter; c) Correct: shark, Predicted: dolphin}
  \label{ige}
\end{figure}

From the ResNet50 and EfficientNet-B0 ignored image samples shown in Figures \ref{igr}\ \& \ref{ige}, we can observe that the samples themselves are confusing. For instance, the lawn\_mower, the boy, snake, seal, and shark images can surely be treated as sweet\_pepper, a girl, worm, otter, and a dolphin, respectively, due to their confusing nature. Hence, the performance of the UQ technique is pretty good not only in terms of the performance metrics(F1-Score, number of ignored samples) but also from a visual aspect.

\textbf{Misclassified (\%) Metric Analysis}
Furthermore, we also analyzed the Misclassified(\%) metric as it is important to see whether or not the ignored samples were confusing and misclassified by the original quantized model. After doing this experiment, we were able to see from Table \ref{cfar100_misclassified} that truly a large portion of these ignored samples are getting misclassified by the original quantized model for both ResNet50(15 samples out of 23) and EfficientNetB0(15 samples out of 19). Hence, this analysis also supports the idea that the UQ technique helps produce usable predictions by ignoring confusing OOD samples.

\begin{table}[htbp]
\caption{Misclassified (\%) Analysis for CIFAR100}
\begin{center}
\begin{tabular}{|c|c|c|}
\hline
\textbf{Model} & \textbf{Misclassified Ignored} & \textbf{Correctly Classified} \\
\textbf{Type} & \textbf{Samples (\%)} & \textbf{Ignored Samples (\%)}\\
\hline
\textbf{ResNet50} & $65.21\%$ & $35.79\%$ \\
\textbf{(23 ignored)} & (15 samples) & (8 samples)\\
\hline
\textbf{EfficientNetB0} & $78.94\%$ & $21.06\%$ \\
\textbf{(19 ignored)} & (15 samples) & (4 samples)\\
\hline
\end{tabular}
\label{cfar100_misclassified}
\end{center}
\end{table}

\subsubsection{CIFAR-100C Experiments}
Apart from the CIFAR-100 dataset, we tested our proposed technique on the CIFAR-100C dataset, where we had a corrupted dataset of different noise types for Severity 1 and Severity 5 and tested on 20 test samples for each noise type and severity.\\

\textbf{Severity-1 Experiments}
In this section, we used Severity-1 test images to make inferences while following the UQ technique and computed performance metrics of F1-Score and the number of Ignored samples(I) for both ResNet50 and EfficientNet-B0 as described in Tables \ref{cfar100c_sev1_resnet50} \& \ref{cfar100c_sev1_efficientnet}. From them, we can infer that the UQ technique indeed was able to increase the F1-score values for most of the noise types, while maintaining a respectable number of ignored samples.
\begin{table}[htbp]
\caption{CIFAR-100C(Severity-1) Results with ResNet50}
\centering
\begin{tabular}{|c|c|c|c|}
\hline
\textbf{Type of} & \textbf{Without UQ} & \textbf{With UQ} & \textbf{With UQ}\\
\textbf{Corruption} & \textbf{(F1-score)} & \textbf{(F1 score)} & \textbf{(\# Ignored Samples)}\\
\hline
\textbf{Brightness} & 15\% & 55.55\% & 11 \\
\hline
\textbf{Motion} & 15\% & 36.36\% & 9 \\
\textbf{Blur} & & &\\
\hline
\textbf{Defocus} & 35\% & 55.55\% & 11 \\
\textbf{Blur} & & &\\
\hline
\textbf{Gaussian} & 25\% & 37.5\% & 12 \\
\textbf{Blur} & & &\\
\hline
\textbf{Glass Blur} & 15\% & 42.85\% & 13 \\
\textbf{Blur} & & &\\
\hline
\textbf{Snow} & 15\% & 9.09\% & 9 \\
\hline
\textbf{Saturate} & 20\% & 33.33\% & 14 \\
\hline
\textbf{Elastic} & 15\% & 33.33\% & 11 \\
\textbf{Transform} & & &\\
\hline
\textbf{Speckle Noise} & 20\% & 0\% & 19 \\
\hline
\textbf{Frost} & 15\% & 22.22\% & 11 \\
\hline
\textbf{Fog} & 20\% & 26.67\% & 5 \\
\hline
\textbf{Jpeg} & 20\% & 28.57\% & 13 \\
\textbf{Compression} & & &\\
\hline
\textbf{Impulse} & 10\% & 100\% & 19 \\
\textbf{Noise} & & &\\
\hline
\textbf{Zoom} & 30\% & 45.45\% & 9 \\
\textbf{Blur} & & &\\
\hline
\textbf{Shot} & 0\% & 0\% & 19 \\
\textbf{Noise} & & &\\
\hline
\textbf{Gaussian} & 0\% & 0\% & 19 \\
\textbf{Noise} & & &\\
\hline
\textbf{Pixelate} & 10\% & 33.33\% & 14 \\
\hline
\textbf{Contrast} & 25\% & 28.57\% & 6 \\
\hline
\end{tabular}
\label{cfar100c_sev1_resnet50}
\end{table}

\begin{table}[htbp]
\caption{CIFAR-100C(Severity-1) Results with EfficientNetB0}
\centering
\begin{tabular}{|c|c|c|c|}
\hline
\textbf{Type of} & \textbf{Without UQ} & \textbf{With UQ} & \textbf{With UQ}\\
\textbf{Corruption} & \textbf{(F1-score)} & \textbf{(F1 score)} & \textbf{(\# Ignored Samples)}\\
\hline
\textbf{Brightness} & 30\% & 46.15\% & 7 \\
\hline
\textbf{Motion} & 30\% & 50\% & 8 \\
\textbf{Blur} & & &\\
\hline
\textbf{Defocus} & 40\% & 54.54\% & 9 \\
\textbf{Blur} & & &\\
\hline
\textbf{Gaussian} & 20\% & 55.55\% & 11 \\
\textbf{Blur} & & &\\
\hline
\textbf{Glass Blur} & 25\% & 33.33\% & 8 \\
\textbf{Blur} & & &\\
\hline
\textbf{Snow} & 20\% & 50\% & 12 \\
\hline
\textbf{Saturate} & 35\% & 33.33\% & 14 \\
\hline
\textbf{Elastic} & 35\% & 54.54\% & 9 \\
\textbf{Transform} & & &\\
\hline
\textbf{Speckle Noise} & 30\% & 80\% & 15 \\
\hline
\textbf{Frost} & 25\% & 55.55\% & 11 \\
\hline
\textbf{Fog} & 35\% & 53.84\% & 7 \\
\hline
\textbf{Jpeg} & 40\% & 33.33\% & 11 \\
\textbf{Compression} & & &\\
\hline
\textbf{Impulse} & 20\% & 75\% & 16 \\
\textbf{Noise} & & &\\
\hline
\textbf{Zoom} & 30\% & 54.54\% & 9 \\
\textbf{Blur} & & &\\
\hline
\textbf{Shot} & 20\% & 50\% & 12 \\
\textbf{Noise} & & &\\
\hline
\textbf{Gaussian} & 0\% & 0\% & 19 \\
\textbf{Noise} & & &\\
\hline
\textbf{Pixelate} & 40\% & 41.67\% & 8 \\
\hline
\textbf{Contrast} & 50\% & 41.67\% & 8 \\
\hline
\end{tabular}
\label{cfar100c_sev1_efficientnet}
\end{table}

Apart from the above experiments, we did the misclassified (\%) metric analysis for 4 of the 19 noises that are part of the CIFAR100C dataset, which are - defocus blur, snow, elastic transform and gaussian noise as described in Tables \ref{cfar100c_sev1_misclassified_resnet50} \& \ref{cfar100c_sev1_misclassified_efficientnetb0} for ResNet50 and EfficientNetB0 respectively. 

\begin{table}[htbp]
\caption{Misclassified (\%) Analysis for CIFAR100C(Severity-1) with ResNet50}
\begin{center}
\begin{tabular}{|c|c|c|c|c|}
\hline
\textbf{Noise Type} & \textbf{Defocus Blur} & \textbf{Snow} & \textbf{Elastic} & \textbf{Gaussian}\\
\textbf{} & \textbf{Blur} & \textbf{} & \textbf{Transform} & \textbf{Noise}\\
\hline
\textbf{Misclassfied (\%)} & $57.14\%$ & $54.54\%$ & $37.50\%$ & $33.30\%$\\
\hline
\end{tabular}
\label{cfar100c_sev1_misclassified_resnet50}
\end{center}
\end{table}

\begin{table}[htbp]
\caption{Misclassified (\%) Analysis for CIFAR100C(Severity-1) with EfficientNetB0}
\begin{center}
\begin{tabular}{|c|c|c|c|c|}
\hline
\textbf{Noise Type} & \textbf{Defocus Blur} & \textbf{Snow} & \textbf{Elastic} & \textbf{Gaussian}\\
\textbf{} & \textbf{Blur} & \textbf{} & \textbf{Transform} & \textbf{Noise}\\
\hline
\textbf{Misclassfied (\%)} & $44.44\%$ & $66.67\%$ & $66.70\%$ & $37.50\%$\\
\hline
\end{tabular}
\label{cfar100c_sev1_misclassified_efficientnetb0}
\end{center}
\end{table}

\begin{figure}[h]
  \includegraphics[width=.15\textwidth]{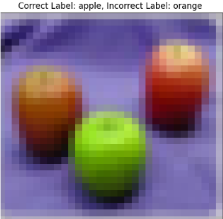}  
  \includegraphics[width=.15\textwidth]{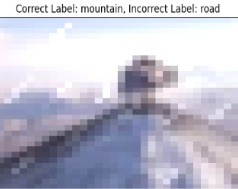}
  \includegraphics[width=.15\textwidth]{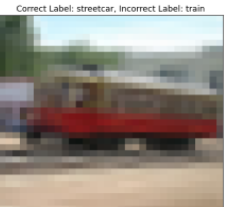}
  \centerline{\includegraphics[width=.15\textwidth]{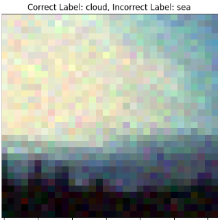}}
  \caption{Some images ignored by ResNet50 for Seversity 1 a) For Defocus Blur - Correct: apple, Predicted: orange; b) For Snow - Correct: mountain, Predicted: road; c) For Elastic Transform - Correct: streetcar, Predicted: train; d) For Gaussian Noise - Correct: cloud, Predicted: sea)}
\label{cifarcsevoneres}
\end{figure}

\begin{figure}[h]
  \includegraphics[width=.15\textwidth]{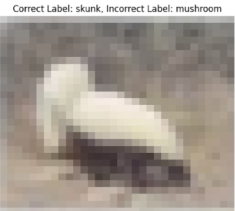}  
  \includegraphics[width=.15\textwidth]{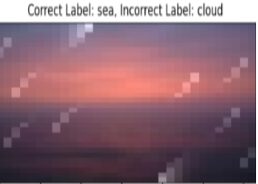}
  \includegraphics[width=.15\textwidth]{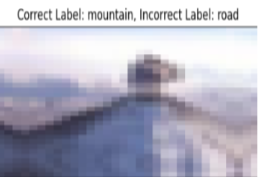}
  \centerline{\includegraphics[width=.15\textwidth]{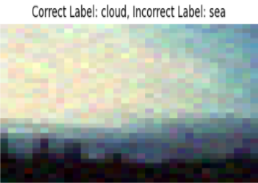}}
  \caption{Some images ignored by EfficientNetB0 for Severity 1 a) For Defocus Blur - Correct: skunk, Predicted: mushroom; b) For Snow - Correct: sea, Predicted: cloud; c) For Elastic Transform - Correct: mountain, Predicted: road; d) For Gaussian Noise - Correct: cloud, Predicted: sea)}
\label{cifarcsevoneeff}
\end{figure}

Lastly, we also did a visual analysis in Figures \ref{cifarcsevoneres} \& \ref{cifarcsevoneeff} of the ignored sample images that the model misclassified and observed that images of apple, mountain, streetcar, cloud, and skunk could be confusing and treated as orange, road, train, sea, and mushroom, respectively. Hence, the above visual analysis also supports the basis of the UQ technique.\\

\textbf{Severity-5 Experiments}\\
We also experimented with a higher degree of severity, i.e., Severity 5, where the noise level was much higher than Severity 1 samples. We followed a similar experimental procedure as we did for Severity 1 samples. 

\begin{table}[htbp]
\caption{CIFAR-100C(Severity-5) Results with ResNet50}
\centering
\begin{tabular}{|c|c|c|c|}
\hline
\textbf{Type of} & \textbf{Without UQ} & \textbf{With UQ} & \textbf{With UQ}\\
\textbf{Corruption} & \textbf{(F1-score)} & \textbf{(F1 score)} & \textbf{(\# Ignored Samples)}\\
\hline
\textbf{Brightness} & 25\% & 50\% & 10 \\
\hline
\textbf{Motion} & 20\% & 23.07\% & 7 \\
\textbf{Blur} & & &\\
\hline
\textbf{Defocus} & 20\% & 44.44\% & 11 \\
\textbf{Blur} & & &\\
\hline
\textbf{Gaussian} & 20\% & 44.44\% & 11 \\
\textbf{Blur} & & &\\
\hline
\textbf{Glass Blur} & 15\% & 30\% & 10 \\
\textbf{Blur} & & &\\
\hline
\textbf{Snow} & 20\% & 28.57\% & 13 \\
\hline
\textbf{Saturate} & 20\% & 60\% & 15 \\
\hline
\textbf{Elastic} & 15\% & 27.27\% & 9 \\
\textbf{Transform} & & &\\
\hline
\textbf{Speckle Noise} & 15\% & 50\% & 18 \\
\hline
\textbf{Frost} & 15\% & 22.22\% & 11 \\
\hline
\textbf{Fog} & 25\% & 23.07\% & 7 \\
\hline
\textbf{Jpeg} & 25\% & 28.57\% & 13 \\
\textbf{Compression} & & &\\
\hline
\textbf{Impulse} & 10\% & 100\% & 19 \\
\textbf{Noise} & & &\\
\hline
\textbf{Zoom} & 25\% & 44.44\% & 11 \\
\textbf{Blur} & & &\\
\hline
\textbf{Shot} & 10\% & 100\% & 19 \\
\textbf{Noise} & & &\\
\hline
\textbf{Gaussian} & 10\% & 0\% & 19 \\
\textbf{Noise} & & &\\
\hline
\textbf{Pixelate} & 15\% & 16.67\% & 14 \\
\hline
\textbf{Contrast} & 10\% & 30\% & 10 \\
\hline
\end{tabular}
\label{cfar100c_sev5_resnet50}
\end{table}

\begin{table}[htbp]
\caption{CIFAR-100C(Severity-5) Results with EfficientNetB0}
\centering
\begin{tabular}{|c|c|c|c|}
\hline
\textbf{Type of} & \textbf{Without UQ} & \textbf{With UQ} & \textbf{With UQ}\\
\textbf{Corruption} & \textbf{(F1-score)} & \textbf{(F1 score)} & \textbf{(\# Ignored Samples)}\\
\hline
\textbf{Brightness} & 15\% & 55.55\% & 11 \\
\hline
\textbf{Motion} & 15\% & 36.36\% & 9 \\
\textbf{Blur} & & &\\
\hline
\textbf{Defocus} & 35\% & 55.55\% & 11 \\
\textbf{Blur} & & &\\
\hline
\textbf{Gaussian} & 25\% & 37.5\% & 12 \\
\textbf{Blur} & & &\\
\hline
\textbf{Glass} & 15\% & 42.85\% & 13 \\
\textbf{Blur} & & &\\
\hline
\textbf{Snow} & 15\% & 9.09\% & 9 \\
\hline
\textbf{Saturate} & 20\% & 33.33\% & 14 \\
\hline
\textbf{Elastic} & 15\% & 33.33\% & 11 \\
\textbf{Transform} & & &\\
\hline
\textbf{Speckle} & 20\% & 0\% & 19 \\
\textbf{Noise} & & &\\
\hline
\textbf{Frost} & 15\% & 22.22\% & 11 \\
\hline
\textbf{Fog} & 20\% & 26.67\% & 5 \\
\hline
\textbf{Jpeg} & 20\% & 28.57\% & 13 \\
\textbf{Compression} & & &\\
\hline
\textbf{Impulse} & 10\% & 100\% & 19 \\
\textbf{Noise} & & &\\
\hline
\textbf{Zoom} & 30\% & 45.45\% & 9 \\
\textbf{Blur} & & &\\
\hline
\textbf{Shot} & 0\% & 0\% & 19 \\
\textbf{Noise} & & &\\
\hline
\textbf{Gaussian} & 0\% & 0\% & 19 \\
\textbf{Noise} & & &\\
\hline
\textbf{Pixelate} & 10\% & 33.33\% & 14 \\
\hline
\textbf{Contrast} & 25\% & 28.57\% & 6 \\
\hline
\end{tabular}
\label{cfar100c_sev5_efficient}
\end{table}

At first, we compared the UQ technique performance with the original quantized model based on F1-Score and the number of Ignored samples for both ResNet50 and EfficientNetB0, as described in Tables \ref{cfar100c_sev5_resnet50} \& \ref{cfar100c_sev5_efficient}. This set of experiments also agrees with Severity 1 results, i.e., the UQ technique improved the F1-score for almost all the noises while keeping a respectable number of ignored samples.

Following this, we used the misclassified(\%) metric as well to see the behavior of the original quantized model on the ignored samples for both ResNet50 and EfficientNetB0. We can observe from Tables \ref{cfar100c_sev5_misclassified_resnet50} \& \ref{cfar100c_sev5_misclassified_efficientnetb0}, that for all the four different noises - defocus blur, snow, elastic transform, and gaussian noise, a good amount of ignored samples are being misclassified by the original quantized model.

\begin{table}[htbp]
\caption{Misclassified (\%) Analysis for CIFAR100C(Severity-5) with ResNet50}
\begin{center}
\begin{tabular}{|c|c|c|c|c|}
\hline
\textbf{Noise Type} & \textbf{Defocus Blur} & \textbf{Snow} & \textbf{Elastic} & \textbf{Gaussian}\\
\textbf{} & \textbf{Blur} & \textbf{} & \textbf{Transform} & \textbf{Noise}\\
\hline
\textbf{Misclassfied (\%)} & $45.45\%$ & $38.46\%$ & $33.33\%$ & $42.10\%$\\
\hline
\end{tabular}
\label{cfar100c_sev5_misclassified_resnet50}
\end{center}
\end{table}

\begin{table}[htbp]
\caption{Misclassified (\%) Analysis for CIFAR100C(Severity-5) with EfficientNetB0}
\begin{center}
\begin{tabular}{|c|c|c|c|c|}
\hline
\textbf{Noise Type} & \textbf{Defocus Blur} & \textbf{Snow} & \textbf{Elastic} & \textbf{Gaussian}\\
\textbf{} & \textbf{Blur} & \textbf{} & \textbf{Transform} & \textbf{Noise}\\
\hline
\textbf{Misclassfied (\%)} & $45.45\%$ & $44.44\%$ & $54.54\%$ & $47.36\%$\\
\hline
\end{tabular}
\label{cfar100c_sev5_misclassified_efficientnetb0}
\end{center}
\end{table}

\begin{figure}[h]
  \includegraphics[width=.15\textwidth]{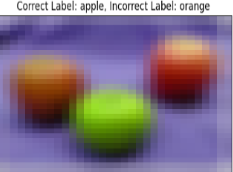}  
  \includegraphics[width=.15\textwidth]{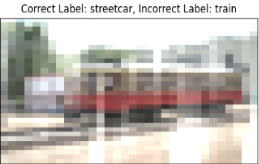}
  \includegraphics[width=.15\textwidth]{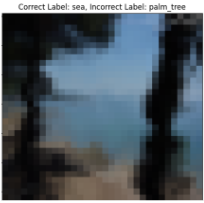}
  \centerline{\includegraphics[width=.15\textwidth]{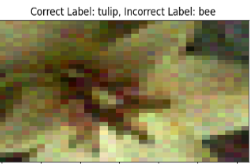}}
  \caption{Some images ignored by ResNet50 for Severity 5 a) For Defocus Blur - Correct: apple, Predicted: orange; b) For Snow - Correct: streetcar, Predicted: train; c) For Elastic Transform 0 Correct: sea, Predicted: palm\_tree; d) For Gaussian Noise - Correct: tulip, Predicted: bee)}
  \label{cifarcsevfiveres}
\end{figure}

\begin{figure}
  \includegraphics[width=.15\textwidth]{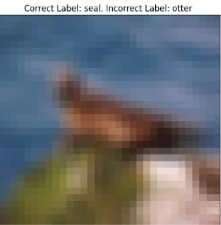}  
  \includegraphics[width=.15\textwidth]{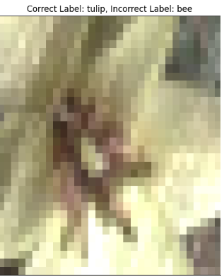}
  \includegraphics[width=.15\textwidth]{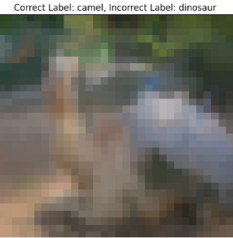}
  \centerline{\includegraphics[width=.15\textwidth]{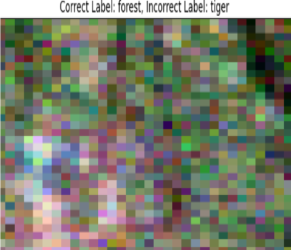}}
  \caption{Some images ignored by EfficientNetB0 for Severity 5 a) For Defocus Blur - Correct: seal, Predicted: otter; b) For Snow - Correct: tulip, Predicted: bee; c) For Elastic Transform - Correct: camel, Predicted: dinosaur; d) For Gaussian Noise - Correct: tiger, Predicted: forest)}
  \label{cifarcsevfiveeff}
\end{figure}

Lastly, we again carried out a visual analysis as in Figures \ref{cifarcsevfiveres} \& \ref{cifarcsevfiveeff} of the ignored sample images that the model misclassified. We again found evidence from this set of examples that the ignored samples are confusing and difficult to classify into 1 class.\\

\subsubsection{Inference Time Comparisons}
Apart from validating the performance of the UQ technique, we also performed an inference time comparison between the UQ technique and the original quantized model. We considered 1 sample for inference and found the time it took for both techniques to provide a prediction, as described in Table \ref{inference_comp}. We observed that the inference time for the UQ technique is 7s in comparison to the 243ms that the original quantized model takes for ResNet50. Similarly, the inference time in the case of EfficientNetB0 also came out as 7.94s for UQ and 196ms for the original model. This analysis certainly shows a drawback in the inference time of the UQ technique.

\begin{table}[htbp]
\caption{Inference time comparisons of UQ technique v/s Original Quantized model}
\begin{center}
\begin{tabular}{|c|c|c|}
\hline
\textbf{Model Type} & \textbf{Quantized Model} & \textbf{UQ Technique} \\
\hline
\textbf{ResNet50} & $243\, \text{ms} \pm 9.54\, \text{ms}$ & $7.09\, \text{s} \pm 1.66\, \text{s}$ \\
\hline
\textbf{EfficientNet B0} & $196\, \text{ms} \pm 8.54\, \text{ms}$ & $7.94\, \text{s} \pm 0.73\, \text{s}$ \\
\hline
\end{tabular}
\label{inference_comp}
\end{center}
\end{table}

\section{Conclusion}
Uncertainty quantification extracts useful predictions while ignoring samples the network cannot classify. The original quantized model misclassified samples that were ignored through this technique. On visual examination, samples ignored constitute very confusing images. Increasing the severity of the noise of images in CIFAR100C increased the number of ignored samples. Quantization helped in reducing the model size by approximately four times.

\section{Future Work}
Estimating uncertainty through frequentist estimation techniques may not be as accurate as through methods such as Bayesian Neural Networks(BNN), which could take into account prior information to better estimate the uncertainty in the model's predictions. However, Such techniques are limited by available computing resources and could not be performed in this study.

When performing inference time dropouts, the computations until the last few layers need to be performed only once as they are static. They can be stored and then multiplied with the dynamic weights of the last few layers over multiple iterations to get the different outputs. This could reduce the time needed to perform uncertainty quantification.

\section{Acknowledgements}
We thank Dr. Jung-Eun Kim(Assistant Professor at NC State University, Raleigh) for her insights to the project during the semester.


\bibliographystyle{IEEEtran}
\bibliography{ref.bib}
\vspace{12pt}
\end{document}